\documentclass[10pt,twocolumn,letterpaper]{article}

\usepackage{iccv}
\usepackage{times}
\usepackage{epsfig}
\usepackage{graphicx}
\usepackage{amsmath}
\usepackage{amssymb}

\usepackage{color}
\usepackage{slashbox}
\usepackage{subfig}

\usepackage[pagebackref=false,breaklinks=true,letterpaper=true,colorlinks,bookmarks=false]{hyperref}

\usepackage{enumitem}

\newcommand*\samethanks[1][\value{footnote}]{\footnotemark[#1]}

\iccvfinalcopy 

\def\iccvPaperID{26} 

\DeclareMathOperator*{\argmin}{arg\,min}
\newcommand{\RR}{\mathbb{R}}

\ificcvfinal\fi
\begin{document}

\title{Love Thy Neighbors: Image Annotation by Exploiting Image Metadata}

\author{
Justin Johnson\thanks{Indicates equal contribution.}~~~~~~~~Lamberto Ballan\samethanks~~~~~~~~Li Fei-Fei\\
Computer Science Department, Stanford University\\
{\tt\small \{jcjohns,lballan,feifeili\}@cs.stanford.edu}\vspace{-15pt}
}


\maketitle


\begin{abstract}
Some images that are difficult to recognize on their own may become more clear in the context of a 
\emph{neighborhood} of related images with similar social-network metadata.
We build on this intuition to improve multilabel image annotation. Our model uses image metadata
nonparametrically to generate neighborhoods of related images using Jaccard similarities, then uses a deep neural 
network to blend visual information from the image and its neighbors.
Prior work typically models image metadata parametrically; in contrast, our nonparametric treatment allows our model 
to perform well even when the vocabulary of metadata changes between training and testing. We perform 
comprehensive experiments on the NUS-WIDE dataset, where we show that our model outperforms state-of-the-art methods 
for multilabel image annotation even when our model is forced to generalize to new types of metadata.
\vspace{-8pt}
\end{abstract}

\section{Introduction}

\begin{figure}[t!]
\centering
    \subfloat[\label{subfig1:tagit}]{%
    \includegraphics[height=0.25\textwidth]{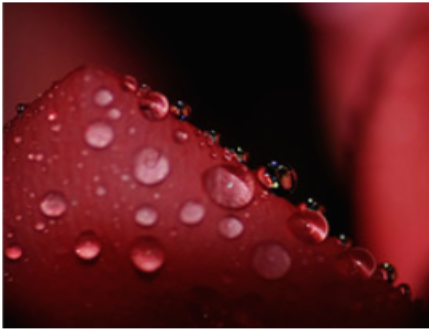}
    }
    \raisebox{0.45mm}{
      \subfloat[\label{subfig2:tagit}]{%
      \includegraphics[height=0.245\textwidth]{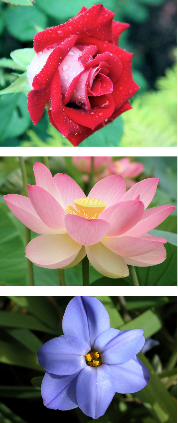}
      }
    }
\vspace{-5pt}
\caption{On its own, the image in (a) is ambiguous - it might be a flower petal, but it could also be a piece of
  fruit or possibly an octopus tentacle. In the context of a \emph{neighborhood} (b) of images with 
  similar metadata, it is more clear that (a) shows a flower. Our model utilizes image neighborhoods to improve multilabel image annotation.}
\vspace{-10pt}  
\label{fig:pullfig}
\end{figure}

Take a look at the image in Figure~\ref{subfig1:tagit}. Might it be a flower petal, or a piece
of fruit, or perhaps even an octopus tentacle? The image on its own is ambiguous. Take another look,
but this time consider that the images in Figure~\ref{subfig2:tagit} share social-network metadata with 
Figure~\ref{subfig1:tagit}. Now the answer is clear: all of these images show flowers.
The context of additional unannotated images disambiguates the visual classification task.
We build on this intuition, showing improvements in multilabel image annotation by exploiting image
metadata to augment each image with a \emph{neighborhood} of related images.

Most images on the web carry metadata; the idea of using it to improve visual classification is not new.
Prior work takes advantage of user tags for image classification and retrieval
\cite{guillaumin-2010,luo-2010,hwang-2012,niu-2014},
uses GPS data \cite{hays-2008,yli-2009,amir-2014} to improve image classification, and utilizes timestamps 
\cite{kim-2010} to both improve recognition and study topical evolution over time.
The motivation behind much of this work is the notion that images with similar metadata
tend to depict similar scenes.

One class of image metadata where this notion is particularly relevant is \emph{social-network metadata}, which can be
harvested for images embedded in social networks such as Flickr. These metadata, such as user-generated
tags and community-curated groups to which an image belongs, are applied to images by people as a means to
communicate with other people; as such, they can be highly informative as to the semantic contents of images.
McAuley and Leskovec~\cite{mcauley-2012} pioneered the study of multilabel image annotation using metadata,
and demonstrated impressive results using only metadata and no visual features whatsoever.

Despite its significance, the applicability of McAuley and Leskovec's method to real-world scenarios is
limited due to the parametric method by which image metadata is modeled. In practice, the vocabulary 
of metadata may shift over time: new tags may become popular, new image groups may be created, etc. An ideal method
should be able to handle such changes, but their method assumes identical vocabularies during training and testing.

In this paper we revisit the problem of multilabel image annotation, taking advantage of both metadata 
and strong visual models. Our key technical contribution is to generate \emph{neighborhoods} of images (as in 
Figure~\ref{fig:pullfig}) nonparametrically using image metadata, then to operate on these 
neighborhoods with a novel parametric model that learns the degree to which visual information from an image and its 
neighbors should be trusted.

In addition to giving state-of-the-art performance on multilabel image annotation (Section~\ref{sec:classification}),
this approach allows our model to perform tasks that are difficult or impossible using existing methods.
Specifically, we show that our model can do the following:
\begin{itemize}
  \setlength\itemsep{1pt}
  \item \textbf{Handle different types of metadata.} We show that the same model can give state-of-the-art
    performance using three different types of metadata (image tags, image sets, and image groups).
    We also show that our model gives strong results when different metadata are available
    at training time and testing time.
  \item \textbf{Adapt to changing vocabularies.} Our nonparametric approach to handling metadata
    allows our model to handle different vocabularies at train and test time. We show that our model gives 
    strong performance even when the training and testing vocabulary of user tags are completely disjoint.
\end{itemize}

\section{Related Work}

\paragraph{Automatic image annotation and image search.}

Our work falls in the broad area of image annotation and search \cite{ssurvey}. Harvesting images from the 
web to train visual classifiers without human annotation is an idea that have been explored many times in the past 
decade \cite{fergus-2005,gwang-2009,feifei-2010,torresani-2010,tsai-2011,neil,levan,xchen-2015}.
Early work on image annotation used voting to transfer labels between visually similar images, often using simple 
nonparametric models \cite{makadia-2008,xli-2009}. 
This strategy is well suited for multimodal data and large vocabularies of weak labels, but is very sensitive to the
metric used to find visual neighbors. Extensions use learnable metrics and weighted voting schemes
\cite{guillaumin-2009,2pknn-2012}, or more carefully select the training images used for voting~\cite{grauman-2014}.
Our method differs from this work because we do not transfer labels from the training set; instead we compute 
nearest-neighbors between \emph{test-set} images using metadata.

These approaches have shown good results, but are limited because they treat tags and visual features separately,
and may be biased towards common labels.
Some authors instead tackle multilabel image annotation by learning parametric models over visual features
that can make predictions \cite{grangier-2008,gwang-2009,zhang-2010,gong-2014} or rank tags \cite{mori-2013}.
Gong \etal \cite{gong-2014} recently showed state of the art results on NUS-WIDE \cite{nuswide} using CNNs with 
multilabel ranking losses. These methods typically do not take advantage of image metadata.
\vspace{-5pt}

\paragraph{Multimodal representation learning: images and tags.}
A common approach for utilizing image metadata is to learn a joint representation of image and tags.
To this end, prior work generatively models the association between visual data and tags or labels 
\cite{lavrenko-2003,barnard-2003,carneiro-2007,srivastava-2012} or applies non-negative matrix factorization to model 
this latent structure \cite{zhu-2010,feng-2014,shah-2014}.
Similarly, Niu \etal \cite{niu-2014} encode the text tags as relations among the images, and define a semi-supervised 
relational topic model for image classification. Another popular approach maps images and tags to a common semantic 
space, using CCA or kCCA \cite{wsabie-2011,hwang-2012,gong-2013,icmr-2014}.
This line of work is closely related to our task, however these approaches only model user tags and assume static 
vocabularies; in contrast we show that our model can generalize to new types of metadata.
\vspace{-8pt}

\paragraph{Beyond images and tags.}
Besides user tags, previous work uses GPS and timestamps \cite{hays-2008,yli-2009,kim-2010,amir-2014} 
to improve classification performance in specific tasks such as landmark classification.
Some authors model the relations between images using multiple metadata 
\cite{stone-2008,mcauley-2012,duan-2014,kumar-2014,fang-2015}.
Duan \etal \cite{duan-2014} present a latent CRF model in which tags, visual features and GPS-tags are used 
jointly for image clustering. McAuley and Leskovec model pairwise social relations between images and then apply a 
structural learning approach for image classification and labeling \cite{mcauley-2012}.
They use this model to analyze the utility of different types of metadata for image labeling. Our work is similarly 
motivated, but their method does not use any visual representation. In contrast, we use a deep neural network to blend 
the visual information of images that share similar metadata.
\vspace{-5pt}



\section{Model}

\begin{figure}
  \centering
  \includegraphics[width=\linewidth]{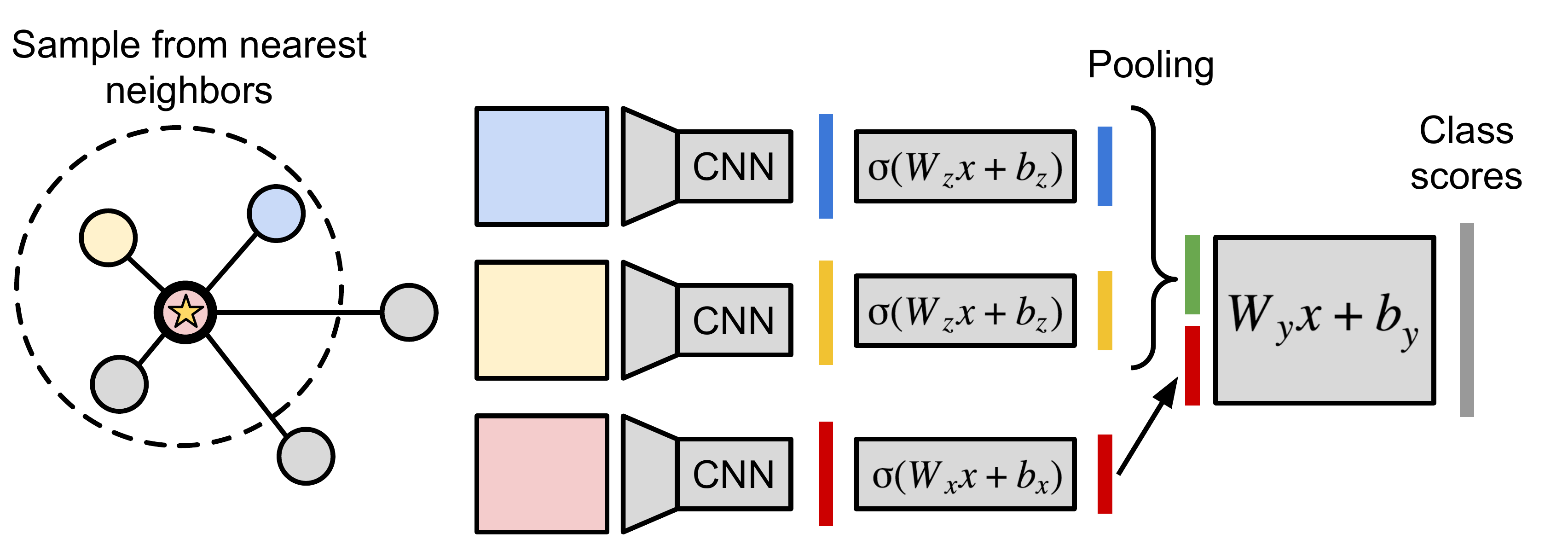}
  \caption{Schematic of our model. To make predictions for an image, we sample several of its nearest neighbors to 
  form a \emph{neighborhood} and we use a CNN to extract visual features. We compute hidden state representations for 
  the image and its neighbors, then operate on the concatenation of these two representations to compute class scores.
  \vspace{-5pt}}
  \label{fig:graph_neighbor_model}
\end{figure}

We design a system that incorporates both visual features of images and the neighborhoods in which they are embedded. 
An ideal system should be able to handle different types of signals, and should be able to generalize to new types of 
image metadata and adapt to their changes over time (e.g. users add new tags or add images to photo-sets).
To this end we use metadata nonparametrically to generate image neighborhoods, then operate on images together with their neighborhoods using a parametric model.
The entire model is summarized in Figure~\ref{fig:graph_neighbor_model}.

Let $X$ be a set of images, $Y$ a set of possible labels, and $\mathcal{D}=\{(x,y)\mid x\in X,\,y\subseteq Y\}$ a 
dataset associating each image with a set of labels.
Let $Z$ be a set of possible neighborhoods for images; in our case a neighborhood is a set of related images, so $Z$ 
is the power set $Z=2^X$.

We use metadata to associate images with neighborhoods. A simple approach would assign each image
$x\in X$ to a single neighborhood $z\in Z$; however there may be more than one useful neighborhood for each image. 
As such, we instead use image metadata to generate a set of \emph{candidate neighborhoods}
$Z_x\subseteq Z$ for each image $x$.

At training time, each element of $Z_x$ is a set of training images, and is computed using training image
metadata. At test time, test image metadata is used to build $Z_x$ from test images; note that
we do not use the training set at test time.

For an image $x\in X$ and neighborhood $z\in Z_x$, we use a function $f$ parameterized by weights $w$
to predict label scores $f(x,z;w)\in\RR^{|Y|}$ for the image $x$. We average these scores over all
candidate neighborhoods for $x$, giving
\begin{equation}
  s(x; w) = \frac{1}{|Z_x|}\sum_{z\in Z_x} f(x, z; w).
\end{equation}
To train the model, we choose a loss $\ell$ and optimize:
\begin{equation}
  w^* = \argmin_w \sum_{(x, y)\in\mathcal{D}} \ell(s(x; w), y).
\end{equation}
The set $Z_x$ may be large, so for computational efficiency we approximate $s(x;w)$ by sampling
from $Z_x$. During training, we draw a single sample during each forward pass and at test time we
use ten samples.

\subsection{Candidate Neighborhoods}
We generate candidate neighborhoods using a nearest-neighbor approach. We use image metadata to compute a distance 
between each pair of images. We fix a \emph{neighborhood size} $m>0$ and a \emph{max rank} $M\geq m$; the candidate 
neighborhoods $Z_x$ for an image $x$ then consist of all subsets of size $m$ of the $M$-nearest neighbors to $x$.

The types of image metadata that we consider are user tags, image photo-sets, and image groups.
Sets are galleries of images collected by the same user (e.g. pictures from the same event such as a wedding).
Image groups are community-curated; images belonging to the same concept, scene or event are uploaded by the social 
network users.
Each type of metadata has a vocabulary $T$ of possible values, and associates each image $x\in X$ with a subset 
$t_x\subseteq T$ of values.
For tags, $T$ is the set of all possible user tags and $t_x$ are the tags for image $x$;
for groups (and sets), $T$ is the set of all groups (sets), and $t_x$ are the groups (sets) to which $x$ belongs.
For sets and groups, we use the entire vocabulary $T$; in the case of tags we follow
\cite{mcauley-2012} and select only the $\tau$ most frequently occurring tags on the training set.

We compute the distance between images using the Jaccard similarity between their image metadata.
Concretely, for $x,x'\in X$ we compute

\vspace{-5pt}
\begin{equation}
  d(x,x') = 1 - |t_x \cap t_{x'}| / |t_x \cup t_{x'}|.
\end{equation}
To prevent an image from appearing in its own neighborhoods, we set $d(x,x)=0$ for all $x\in X$.

Generating candidate neighborhoods introduces several hyperparameters, namely the neighborhood size $m$,
the max rank $M$, the type of metadata used to compute distances, and the tag vocabulary size $\tau$.
We show in Section~\ref{sec:hyperparameters} that the type of metadata is the only hyperparameter that significantly 
affects our performance.

\begin{figure*}
  \centering
  \includegraphics[width=0.92\linewidth]{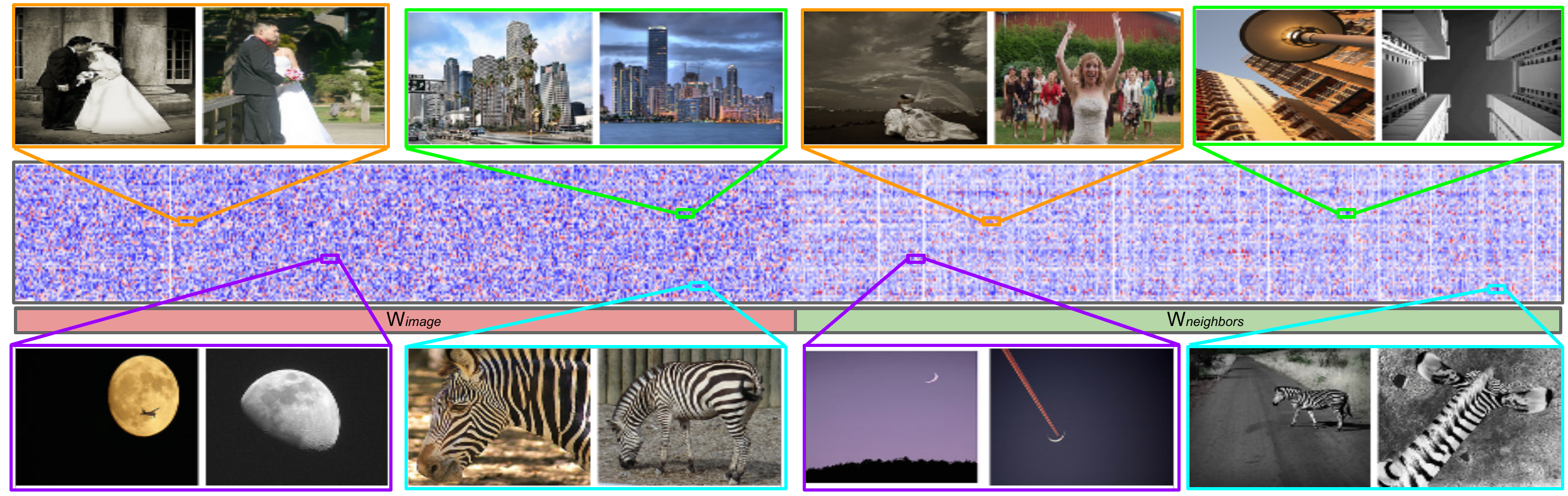}
  \vspace{-5pt}
  \caption{Learned weights $W_y$. The model uses features from both the image and its neighbors. We show examples of 
  images whose label scores are influenced more by the image and by its neighborhood; images with the same 
  ground-truth labels are highlighted with the same colors. Images that are influenced by their neighbors tend to be 
  non-canonical views.\vspace{-8pt}}
  \label{fig:weights}
\end{figure*}

\subsection{Label Prediction} 
Given an image $x\in X$ and a neighborhood $z=\{z_1,\ldots,z_m\}\in Z$, we design a model that incorporates visual 
information from both the image and its neighborhood in order to make predictions for the image. Our model is 
essentially a fully-connected two layer neural network applied to features from the image and its neighborhood, except 
that we pool over the hidden states for the neighborhood images.

We use a CNN \cite{lecun-1998,alex-2012} $\phi$ to extract $d$-dimensional features from the images $x$ and $z_i$.
We compute an $h$-dimensional hidden state for each image by applying an affine transform
and an elementwise ReLU nonlinearity $\sigma(\xi)=\max(0,\xi)$ to its features.
To let the model treat hidden states for the image and its neighborhood differently,
we apply distinct transforms to $\phi(x)$ and $\phi(z_i)$, parameterized by $W_x\in\RR^{d\times h},b_x\in\RR^h$ and
$W_z\in\RR^{d\times h},b_z\in\RR^h$.

At this point we have hidden states $v_x,v_{z_i}\in\RR^h$ for $x$ and each $z_i\in z$;
to generate a single hidden state $v_z\in\RR^h$ for the neighborhood $z$ we pool each $v_{z_i}$
elementwise so that $(v_z)_j = \max_i (v_{z_i})_j$.
Finally to compute label scores $f(x,z;w)\in\RR^{|Y|}$ we concatenate $v_x$
and $v_z$ and pass them through a third affine transform parameterized by
$W_y\in\RR^{2h\times|Y|},b_y\in\RR^{|Y|}$. To summarize:

\vspace{-12pt}
\begin{align}
  v_x &= \sigma(W_x\phi(x) + b_x) \\
  v_z &= \max_{i=1,\ldots,m}\bigg(\sigma(W_z\phi(z_i) + b_z)\bigg) \\
f(x, w; z) &= W_y \begin{bmatrix} v_x \\ v_z \end{bmatrix} + b_y \label{eq:scores}
\end{align}
The learnable parameters are $W_x$, $b_x$, $W_z$, $b_z$, $W_y$, and $b_y$.

\subsection{Learned Weights}\vspace{-5pt}
An example of a learned matrix $W_y$ is visualized in Figure~\ref{fig:weights}.
The left and right sides multiply the hidden states for the image and its neighborhood respectively.
Both sides contain many nonzero weights, indicating that the model learns to use information
from both the image and its neighborhood; however the darker coloration on the left suggests that information
from the image is weighted more heavily.

We can follow this idea further, and use Equation~\ref{eq:scores} to compute for each image the portion of
its score for each label that is due to the hidden state of the image $v_x$ and its neighborhood $v_z$.
The left side of Figure~\ref{fig:weights} shows examples of correctly labeled images whose scores are
more due to the image, while the right shows images more influenced by their neighborhoods.
The former show canonical views (such as a bride and groom for \emph{wedding}) while the latter are
more non-canonical (such as a \emph{zebra} crossing a road).

\subsection{Implementation details}\vspace{-5pt}
We apply $L_2$ regularization to the matrices $W_x,W_z,$ and $W_y$ and apply dropout \cite{hinton2012improving} with 
$p=0.5$ to the hidden layers $h_x$ and $h_z$. We initialize all parameters using the method of \cite{he2015delving} 
and optimize using stochastic gradient descent with a fixed learning rate, RMSProp \cite{tieleman2012lecture}, and a 
minibatch size of 50.
We train all models for 10 epochs, keeping the model snapshot that performs the best on the validation set.
For all experiments we use a learning rate of $1\times10^{-4}$, $L_2$ regularization strength $3\times10^{-3}$
and hidden dimension $h=500$; these values were chosen using grid search.

Our image feature function $\phi$ returns the activations of the last fully-connected layer of the BLVC Reference 
CaffeNet \cite{jia2014caffe}, which is similar to the network architecture of \cite{alex-2012}.
We ran preliminary experiments using features from the model of VGG \cite{vgg-2015}, but this did not significantly 
change the performance of our model.
For all models our loss function $\ell$ is a sum of independent one-vs-all logistic classifiers.

\section{Experimental Protocol}\vspace{-3pt}
\subsection{Dataset}\vspace{-3pt}
In all experiments we use the NUS-WIDE dataset~\cite{nuswide}, which has been widely used for image labeling and 
retrieval. It consists of 269,648 images collected from Flickr, each manually annotated for the presence or absence of 
81 labels. Following \cite{mcauley-2012} we augment the images with metadata using the Flickr API, discarding images 
for which metadata is unavailable.
Following \cite{gong-2014} we also discard images for which all labels are absent. This leaves 
190,253 images, which we randomly partition into training, validation, and test sets of 110K, 40K, and 40,253
images respectively. We generate 5 such splits of the data and run all experiments on all splits. Statistics of
the dataset can be found in Table~\ref{tab:nuswide}. We will make our data and features publicly available to
facilitate future comparisons.

\newlength{\panelwidth}
\setlength{\panelwidth}{0.205\textwidth}
\newlength{\panelhspace}
\setlength{\panelhspace}{-2.75mm}
\begin{figure*}
  \centering
  \includegraphics[width=\panelwidth]{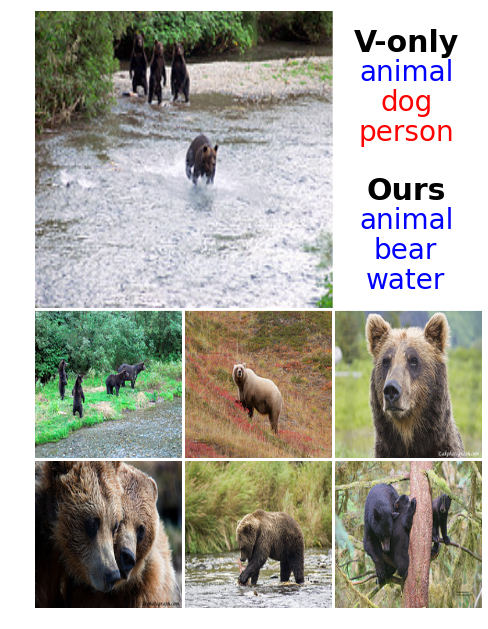}
  \hspace{\panelhspace}
  \includegraphics[width=\panelwidth]{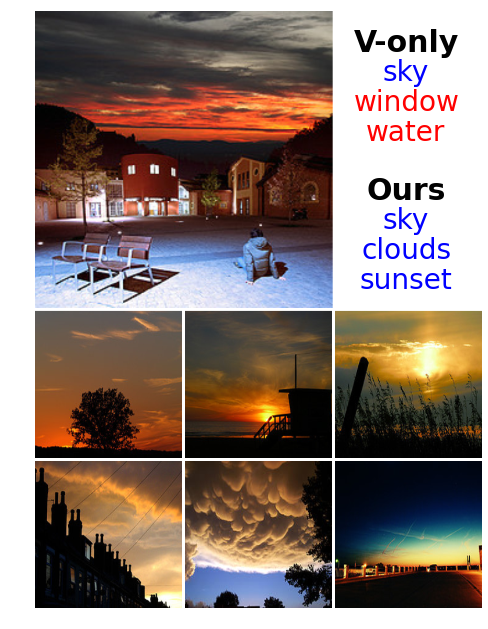}
  \hspace{\panelhspace}
  \includegraphics[width=\panelwidth]{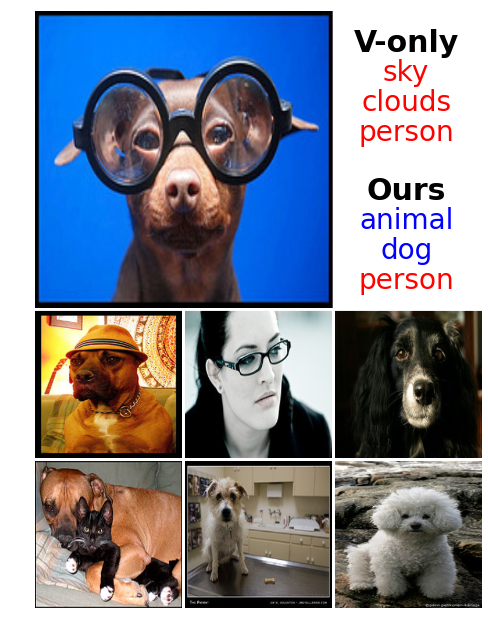}
  \hspace{\panelhspace}
  \includegraphics[width=\panelwidth]{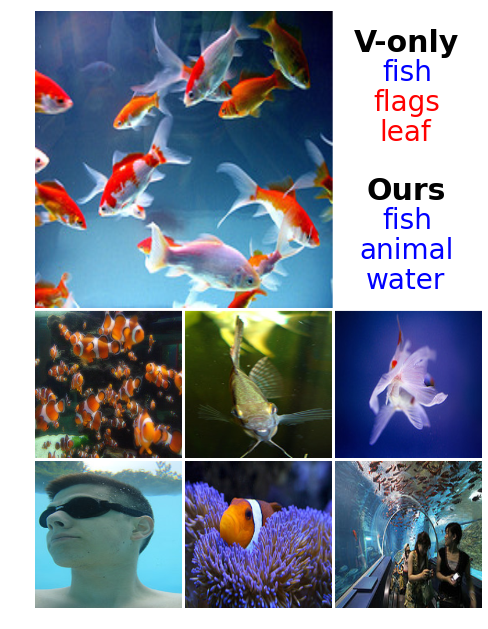}
  \hspace{\panelhspace}
  \includegraphics[width=\panelwidth]{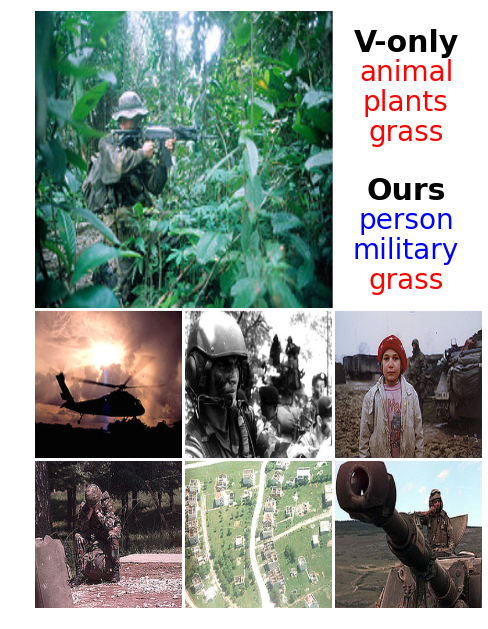} \\
  \vspace{-1mm}
  \includegraphics[width=\panelwidth]{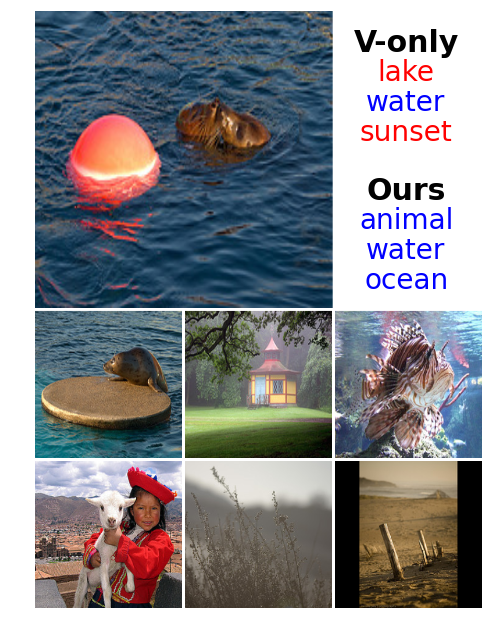}
  \hspace{\panelhspace}
  \includegraphics[width=\panelwidth]{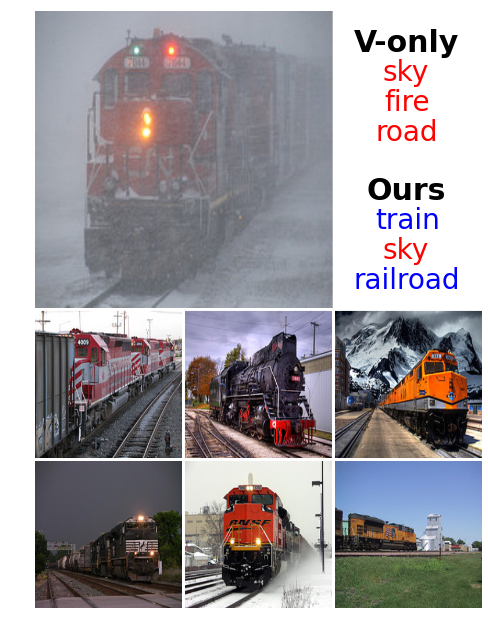}
  \hspace{\panelhspace}
  \includegraphics[width=\panelwidth]{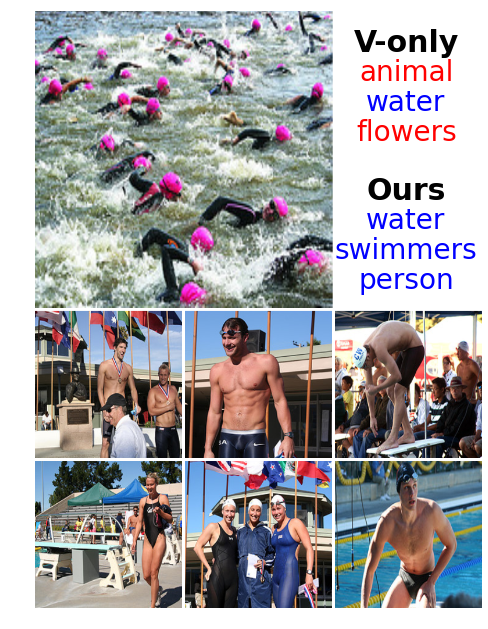}
  \hspace{\panelhspace}
  \includegraphics[width=\panelwidth]{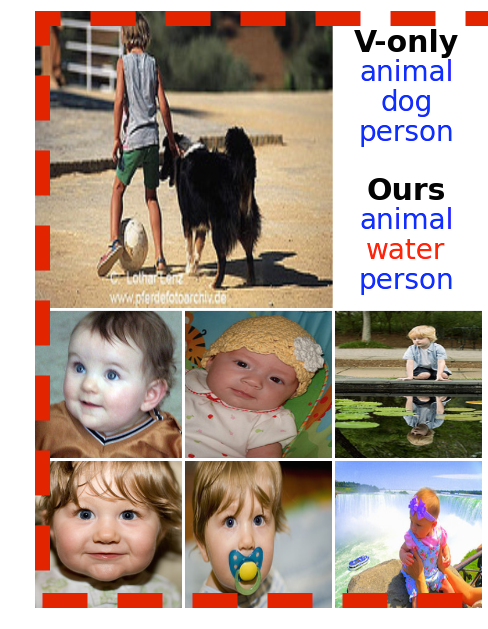}
  \hspace{\panelhspace}
  \includegraphics[width=\panelwidth]{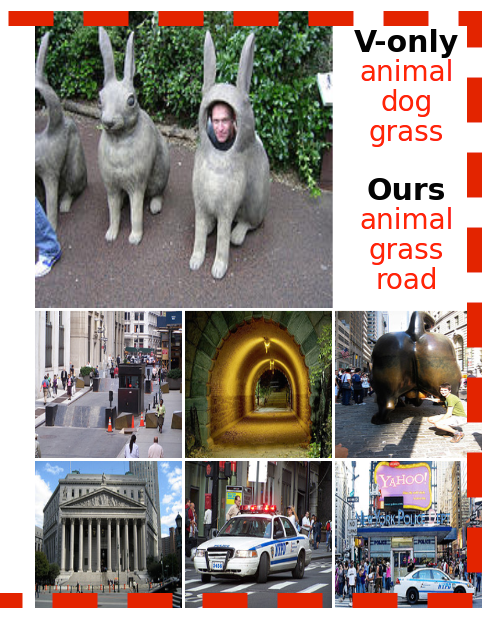}  
  \caption{
    Example results. For each image we show the top 3 scoring labels using the visual-only (V-only) model
    and our model using tag nearest neighbors; correct labels are shown in blue and incorrect labels in red. We also 
    show the 6 nearest neighbors to each image; its neighborhoods are drawn from these images. The red dashed lines 
    show failure cases.\vspace{-5pt}}
  \label{fig:labeling-examples}
\end{figure*}

\vspace{-8pt}
\begin{table}[!h]
  \centering
  \resizebox{0.45\textwidth}{!}{
    \begin{tabular}{c|c|ccc}
      NUS-WIDE & Labels & Tags & Sets & Groups\\
      \hline
      \# unique elements & $81$ & $10,000$ & $165,039$ & $95,358$\\
      \# image per ($.$) & $5701.3$ / $1682$ & $270.3$ / $91$ & $2.3$ / $1$ & $26.1$ / $2$ \\
      \# ($.$) per image & $2.4$ / $2$ & $14.2$ / $11$ & $2.0$ / $1$ & $13.1$ / $8$ \\
      \hline
    \end{tabular}
  }
  \vspace{-5pt}
  \caption{Dataset statistics. Image and ($.$) counts are reported in the format mean / median.}
  \label{tab:nuswide}
\end{table}

\begin{table*}[!th]
\centering
\resizebox{1\textwidth}{!}{
\begin{tabular}{l||cc|ccccc}
Method & mAP$_{L}$ & mAP$_{I}$ & $Rec_{L}$ & $Prec_{L}$ & $Rec_{I}$ & $Prec_{I}$ \\
\hline
\hline
Tag-only Model + linear SVM \cite{mcauley-2012} & 46.67 & - & - & - & - & - \\
Graphical Model (all metadata) \cite{mcauley-2012} & 49.00 & - & - & - & - & - \\
CNN + softmax \cite{gong-2014} & - & - & 31.22 & 31.68 & 59.52 & 47.82 \\
CNN + ranking \cite{gong-2014} & - & - & 26.83 & 31.93 & 58.00 & 46.59 \\
CNN + WARP \cite{gong-2014}    & - & - & 35.60 & 31.65 & 60.49 & 48.59 \\
\hline
\hline
Upper bound & $100.00 \scriptstyle{\pm 0.00}$ & $100.00 \scriptstyle{\pm 0.00}$ & $68.52 \scriptstyle{\pm 0.35}$ & 
$60.68 \scriptstyle{\pm 1.32}$ & $92.09 \scriptstyle{\pm 0.10}$ & $66.83 \scriptstyle{\pm 0.12}$ \\
\hline
Tag-only + logistic & $43.88 \scriptstyle{\pm 0.32}$ & $77.06 \scriptstyle{\pm 0.14}$ & $47.52
\scriptstyle{\pm 2.59}$ & $46.83 \scriptstyle{\pm 0.89}$ & $71.34 \scriptstyle{\pm 0.16}$ & $51.18 \scriptstyle{\pm 
0.16}$ \\
CNN \cite{alex-2012} + kNN-voting \cite{makadia-2008} & $44.03 \scriptstyle{\pm 0.26}$ & $73.72 \scriptstyle{\pm 0.10}$
& $30.83 \scriptstyle{\pm 0.37}$ & $44.41 \scriptstyle{\pm 1.05}$ & $68.06 \scriptstyle{\pm 0.15}$ & $49.49 
\scriptstyle{\pm 0.11}$ \\
CNN \cite{alex-2012} + logistic (visual-only) & $45.78 \scriptstyle{\pm 0.18}$ & $77.15 \scriptstyle{\pm 0.11}$ & 
$43.12 \scriptstyle{\pm 0.39}$ & $40.90 \scriptstyle{\pm 0.39}$ & $71.60 \scriptstyle{\pm 0.19}$ & $51.56 
\scriptstyle{\pm 0.11}$ \\
Image neighborhoods + CNN-voting & $50.40 \scriptstyle{\pm 0.23}$ & $77.86 \scriptstyle{\pm 0.15}$ & 
$34.52 \scriptstyle{\pm 0.47}$ & $\mathbf{56.05} \scriptstyle{\pm 1.47}$ & $72.12 \scriptstyle{\pm 0.21}$ & $51.91
\scriptstyle{\pm 0.20}$ \\
\hline
Our model: tag neighbors& $52.78 \scriptstyle{\pm 0.34}$ & $\mathbf{80.34} \scriptstyle{\pm 0.07}$ & $43.61 
\scriptstyle{\pm 0.47}$ & $46.98 \scriptstyle{\pm 1.01}$ & $74.72 \scriptstyle{\pm 0.16}$ & 
$\mathbf{53.69} \scriptstyle{\pm 0.13}$\\
Our model: tag neighbors + tag vector & $\mathbf{61.88} \scriptstyle{\pm 0.36}$ & $80.27 \scriptstyle{\pm 0.08}$ & 
$\mathbf{57.30} \scriptstyle{\pm 0.44}$ & $54.74 \scriptstyle{\pm 0.63}$ & $\mathbf{75.10} \scriptstyle{\pm 0.20}$ & 
$53.46 \scriptstyle{\pm 0.09}$\\
\hline
\end{tabular}
}
\vspace{-8pt}
\caption{Results on NUS-WIDE. Precision and recall are measured using $n=3$ labels per image. Metrics are reported 
both per-label (mAP$_L$) and per-image (mAP$_I$). We run on 5 splits of the data and report mean and standard 
deviation.}
\vspace{-14pt}  
\label{tab:baselines}
\end{table*}
\vspace{-5mm}

\subsection{Metrics}\label{metrics}\vspace{-5pt}
Prior work uses a variety of metrics and experimental setups on NUS-WIDE, making direct comparisons of results 
difficult. Following prior work \cite{makadia-2008,guillaumin-2009,2pknn-2012,gong-2014} we assign a fixed number of 
labels to each image and report (overall) precision $Prec_I$ and recall $Rec_I$; we also compute the precision and 
recall for each label and report the mean across labels as the \emph{per-label} metrics $Prec_L, Rec_L$.

NUS-WIDE has a highly uneven distribution of labels; the most common (\emph{sky}) has over 68,000 examples
and the least common (\emph{map}) has only 53. As a result the overall precision and recall statistics are
strongly biased towards the common labels. The precision and recall for uncommon labels are extremely noisy
since they are based on only a handful of test-set examples, and the mean per-label statistics inherit this
noise since they weight all classes equally.

Mean Average Precision (mAP) is another widely used metric~\cite{mcauley-2012,ssurvey};
it directly measures ranking quality, so it naturally handles multiple labels
and does not require choosing a fixed number of labels per image.
As with other metrics, we report mAP both per-label (mAP$_L$) and per-image (mAP$_I$).
mAP$_L$ is less noisy and hence preferable to other per-label metrics since it considers the full ranking
of images instead of only the top labels for each image.
\vspace{-5pt}

\section{Experiments}\vspace{-5pt}
\subsection{Multilabel Image Annotation}\label{sec:classification}\vspace{-5pt}
We show that our model achieves state-of-the art results for multilabel image annotation on NUS-WIDE. Our best model 
computes neighborhoods using tags with a vocabulary size of $\tau=5000$, neighborhood size $m=3$ and max-rank $M=6$.
Preliminary experiments at combining all types of metadata did not show improvements over using tags alone.
We also show the result of augmenting the hidden state of our model with a binary indicator vector of image tags.
All results are shown in Table~\ref{tab:baselines}.

\vspace{-7pt}\paragraph{Baselines.}

First we report the results of McAuley and Leskovec~\cite{mcauley-2012} and Gong \etal~\cite{gong-2014} as in their 
original papers. Then we compare our model with four baselines:

1. Tag-only + logistic: the tag-only model of \cite{mcauley-2012} represents each image with a sparse binary vector 
indicating its tags, while their full model uses all available metadata (tags, groups, galleries, and sets) and 
incorporates a graphical model to model pairwise interactions between these features. Unfortunately these results are 
not directly comparable to ours, since they do not discard images without ground-truth labels; as a result they use 
244K images for their experiments while we use only 190K. We reimplement a version of their tag-only model by training 
one-vs-all logistic classifiers on top of binary tag indicator features. Our reimplementation performs slightly worse 
than their reported numbers due to the difference in dataset size.

2. CNN + logistic loss: the results of \cite{gong-2014} have been obtained using a deep convolutional neural 
networks in the style of \cite{alex-2012} equipped with various multilabel loss functions. Again, these results are 
not directly comparable to ours because they train their networks from scratch on the NUS-WIDE dataset, while we use 
networks that were pretrained on ImageNet \cite{imagenet}. We reimplement a version of their model by training 
one-vs-all logistic classifiers using the features extracted from our pretrained network. This is an extremely strong 
baseline; note that it already outperforms \cite{gong-2014}, highlighting the power of the pretrained network.

3. CNN + kNN voting: as an additional baseline we implement a simple nearest neighbor approach.
For each test image we compute the $L_2$ distance between its CNN features and the features of all images in the 
training set; the ground-truth labels of the retrieved training images are then used in a voting scheme similar to 
\cite{makadia-2008,xli-2009}.

4. Image neighborhoods + CNN-voting: for each test image we compute its $M$-nearest neighbors on the 
test set using user tags as in our full model, but instead of passing these neighbors to our parametric model we apply
the CNN+logistic visual-only model to the image and its neighbors. Then we set the label scores of the test image to 
be a weighted sum of its visual-only label scores and the mean of the visual-only label scores of its neighbors.

\begin{figure}
    \centering 
    \subfloat[Example PR curves\label{subfig1:aps}]{%
    \includegraphics[width=0.45\linewidth]{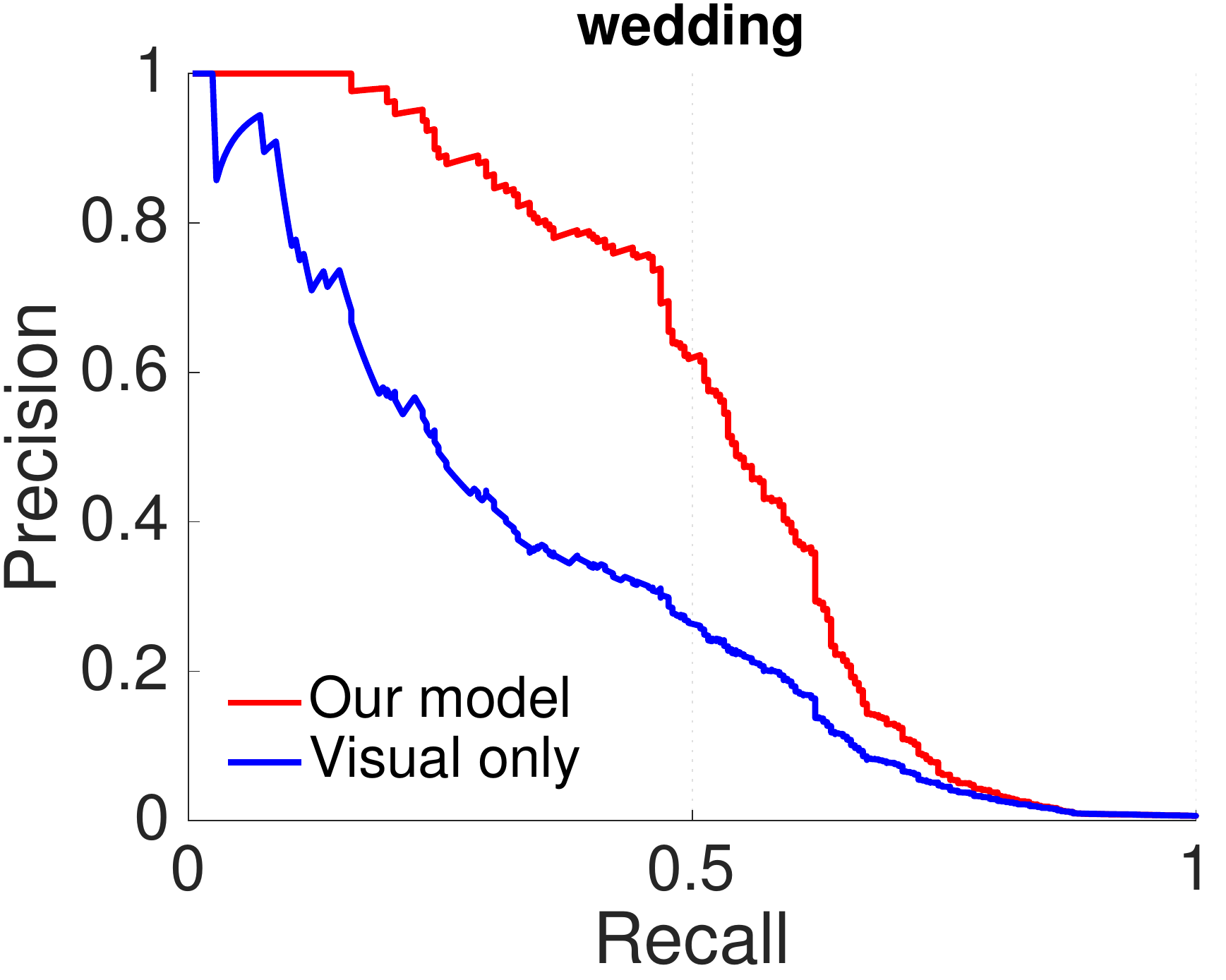} 
    \includegraphics[width=0.45\linewidth]{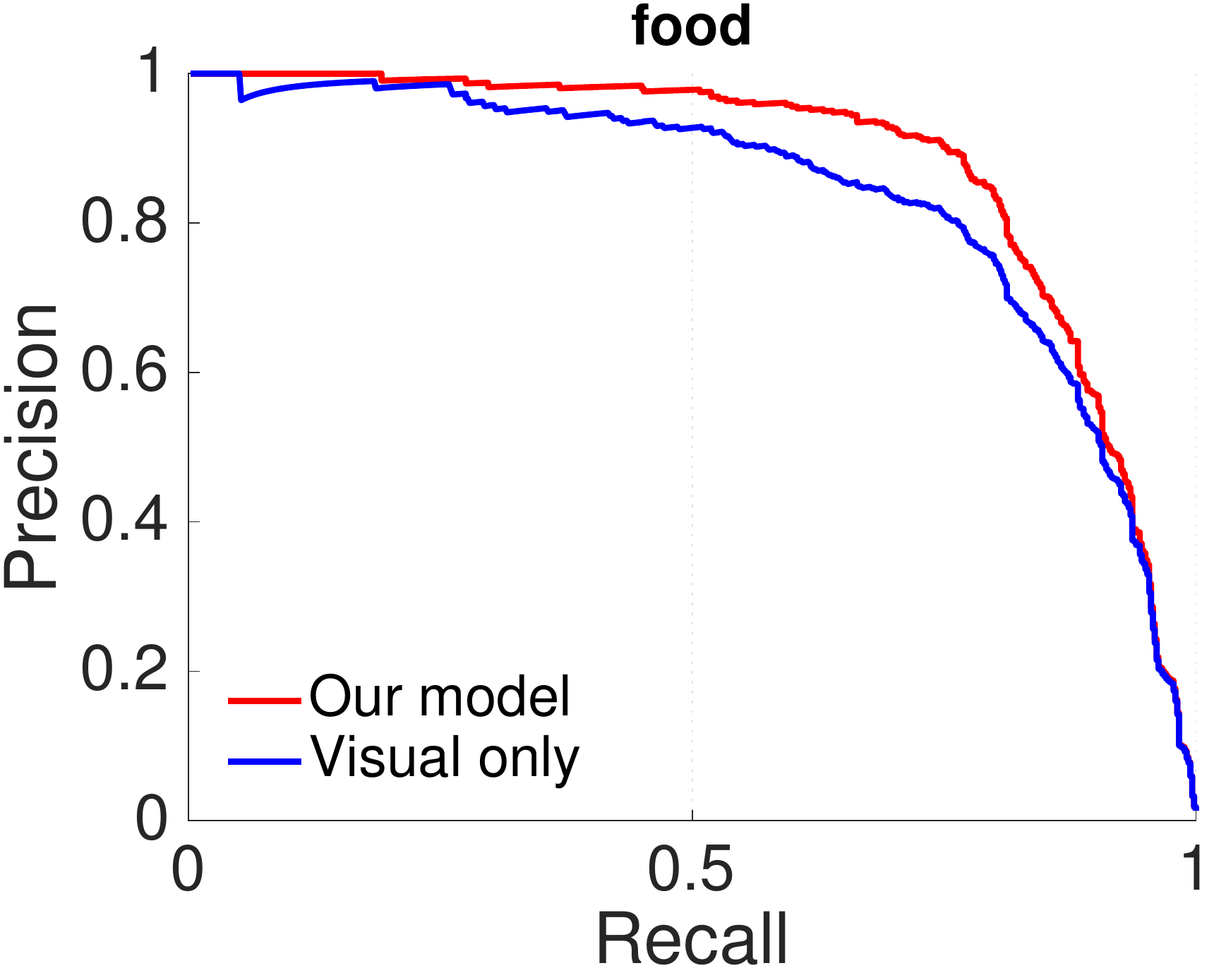}
    }\\\vspace{-3mm}
    \subfloat[Our model \emph{vs} Visual-only model\label{subfig2:aps}]{%
    \includegraphics[width=0.45\linewidth]{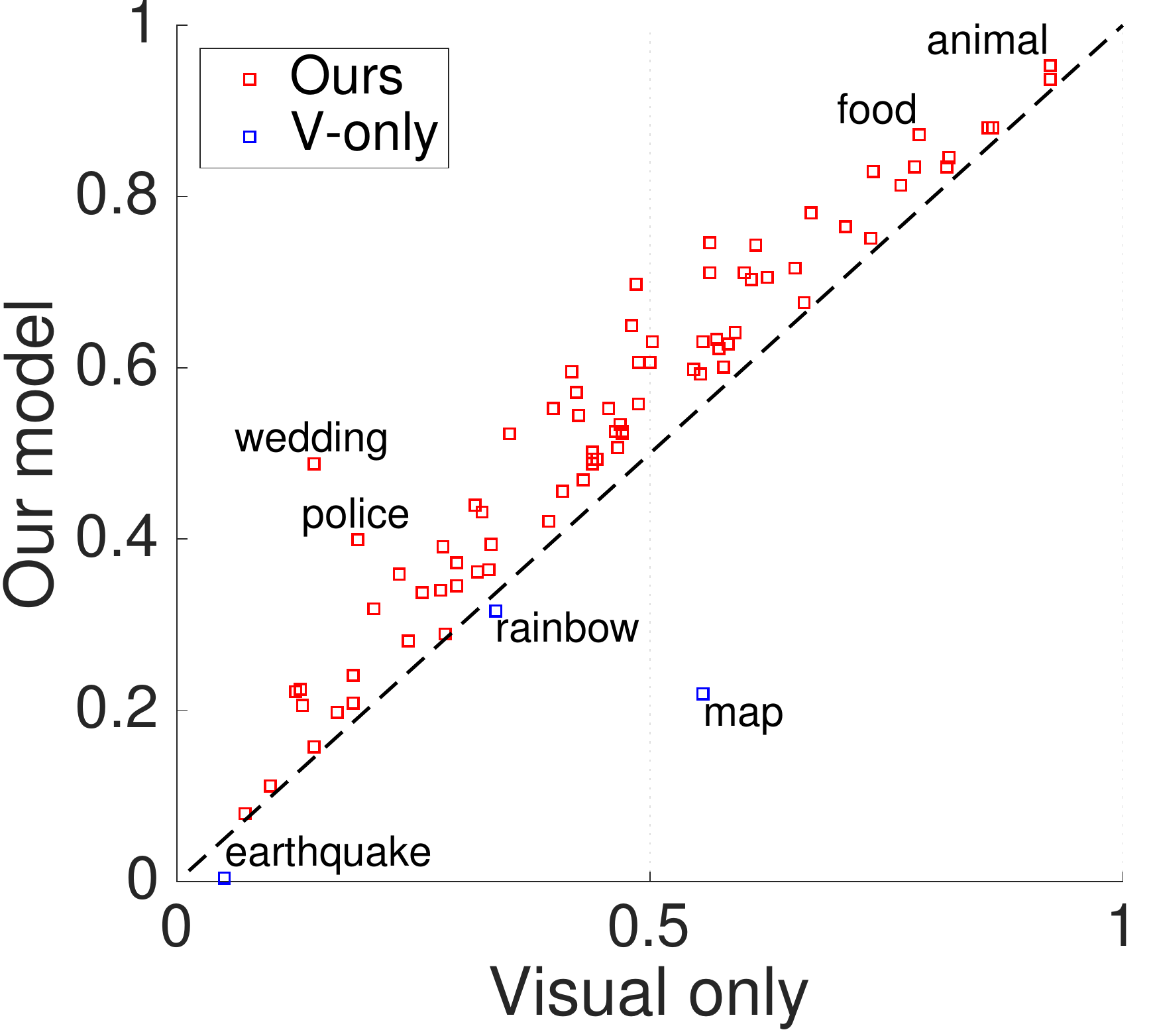} 
    \includegraphics[width=0.45\linewidth]{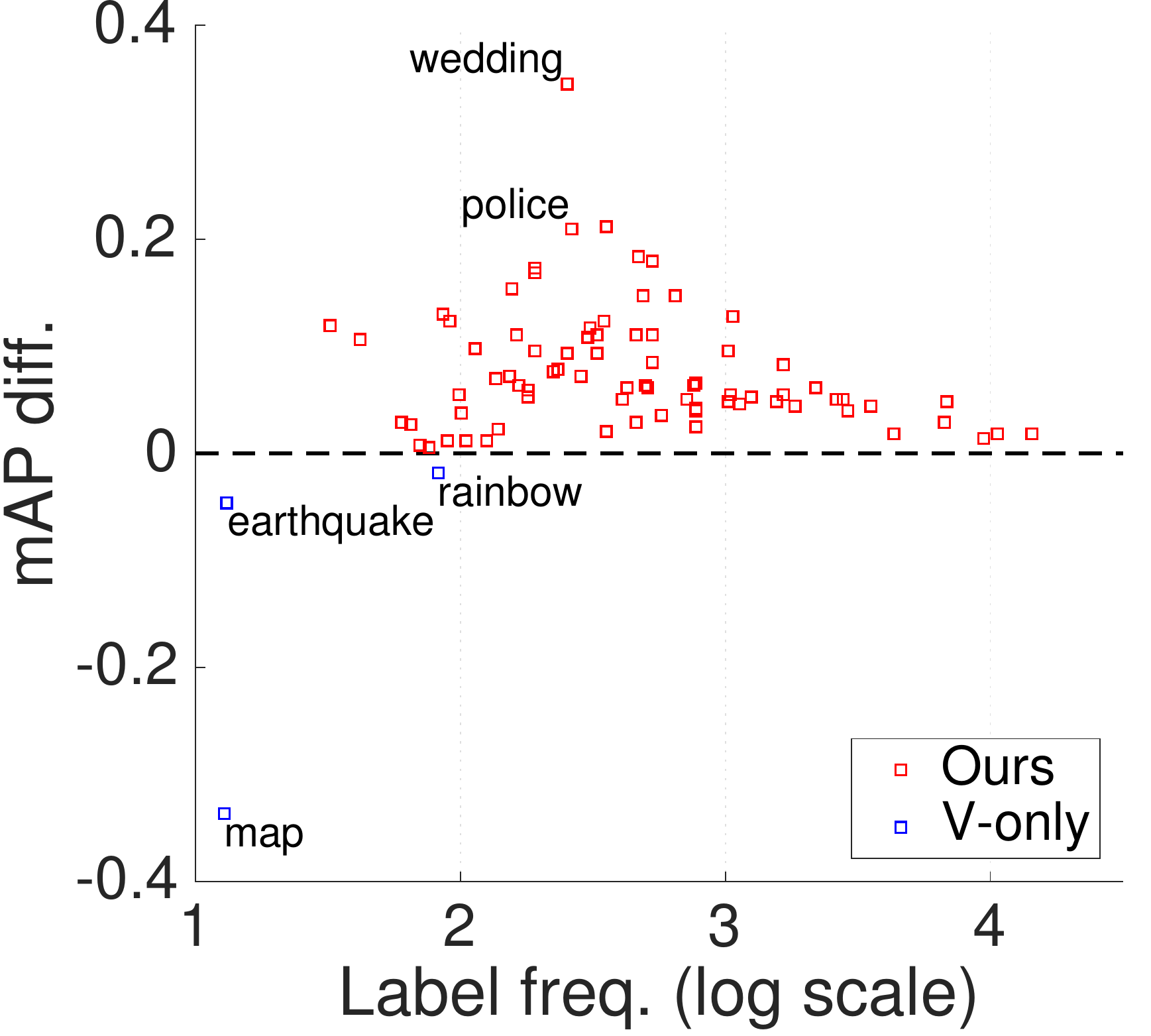}  
    }
    \vspace{-1mm}
    \caption{(a)~Our model shows large improvements for labels with high intra-class variability (e.g. \emph{wedding})
    and for labels where the visual model performs well (e.g. \emph{food}).
    (b)~Left: AP for each label of our model \emph{vs} the baseline; we improve for all but three labels (\emph{map}, 
    \emph{earthquake}, \emph{rainbow}). (b)~Right: difference in class AP between our model and the visual-only model 
    vs label frequency.}
\vspace{-3mm}
\label{fig:pr_class_aps}
\end{figure}

\vspace{-4mm}\paragraph{Upper bound.}
As discussed in Section~\ref{metrics}, we assign the top $n=3$ labels to each image and 
report precision both per-class and per-image (recall that the average number of labels per image is approximately 
$2.4$). However many images do not have exactly 3 ground-truth labels; this means that no classifier can achieve unit 
precision and recall.
To estimate upper bounds for these metrics, we train one-vs-all logistic classifiers where each image
its represented by a binary indicator vector encoding its ground-truth labels. As seen in Table~\ref{tab:baselines},
even this perfect classifier achieves far from perfect performance on many of the evaluation metrics.

\vspace{-4mm}\paragraph{Results.}
Table~\ref{tab:baselines} shows that our model outperforms prior work on nearly all metrics.
The per-class precision and recall metrics display high variance; as a result we do not believe them to be the best 
indicators of performance.
The mAP metrics give a clearer picture of performance, since they display lower variance and do not rely on annotating 
each image with a fixed number of labels. On these metrics our model outperforms all baselines by a significant margin.

As an extension, we append the binary tag vector to the representation learned by our model (tag neighbors + tag 
vector); this does not significantly change performance as measured by per-image metrics, but does
show improvement on per-class metrics. This suggests that the binary tag vector is especially useful for rare
classes which may have strong correlations with certain user tags. Although it increases per-class performance, this 
extension significantly increases the number of learnable parameters and makes generalization to new types of 
metadata impossible.

In order to qualitatively understand some of the cases where our model outperforms the baselines,
Figure~\ref{fig:labeling-examples} compares the top three labels produced by our model and by the visual-only
baseline. The additional visual information provided by the neighborhoods can help resolve ambiguities in
non-canonical views; for example in the image of swimmers the visual-only model appears to mistake the 
colorful swim caps for flowers, but the neighborhood provides canonical views of swimmers.

In few cases the neighborhood can hurt performance. For example in the image of the boy with a dog, the visual-only 
model correctly produces a \emph{dog} label but our model replaces this with a \emph{water} label, likely because no 
neighbors contain dogs but two neighbors contain visible bodies of water.
However the aggregate metrics of Table~\ref{tab:baselines} make it clear that neighborhoods are beneficial more often 
than not. 

There are cases where both models fail; for example see the lower right image of Figure~\ref{fig:labeling-examples}
which shows a person crouching inside a statue of a rabbit. The ground-truth labels for this challenging image are
\emph{statue} and \emph{person}, which are produced by neither model.

\begin{figure}
  \centering
  \includegraphics[width=0.90\linewidth]{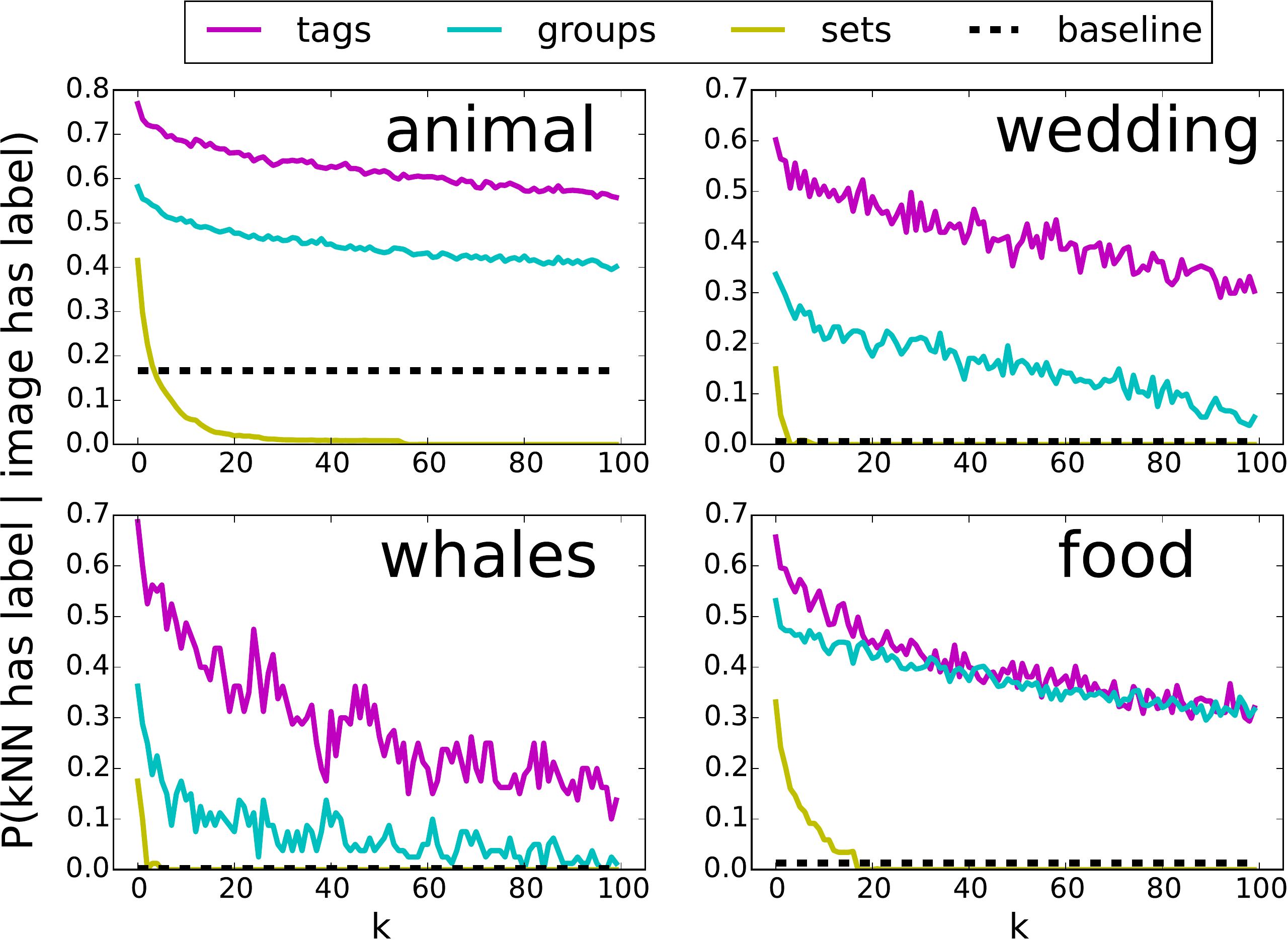}
  \vspace{-5pt}
  \caption{Probability that the $k$th nearest neighbor of an image has a particular label given that the image has the 
  label, as a function of $k$ and using different metadata. The dashed lines give the overall probability that an
  image has the label. Across all metadata and all classes, an image and its neighbors are likely to share labels.
  \vspace{-5pt}}
  \label{fig:prob_vs_k}
  \vspace{-2pt}
\end{figure}

More quantitatively, Figure~\ref{subfig2:aps} compares the average precision (AP) of both our model and the
visual-only baseline for each label; our model outperforms the baseline on all but three labels:
\emph{map}, \emph{earthquake}, and \emph{rainbow}. Of these, \emph{map} is the only label where our model is
significantly outperformed by the baseline. 
Figure~\ref{subfig2:aps} also reveals that these three labels are among the most infrequent; they have only 53, 
56, and 397 instances respectively in the entire dataset, and an average of only 12.8, 13.2, and 82.0 instances 
respectively on the test sets. With so few test instances the performance of both models on these labels is highly 
susceptible to noise.
It is also interesting to note that the middle frequencies are the ones in which our model gives the major boost in 
performance, while for the very frequent labels it is still able to give slight improvements. 

Figure~\ref{subfig1:aps} also shows two example precision-recall curves. 
The \emph{wedding} label has high intra-class variability, making it difficult to recognize using visual features
alone; our model is able to give a large boost in performance by taking advantage of image metadata.
Our model also gives improvements on labels such as \emph{food} where the performance of the visual-only baseline is 
already quite strong.

\subsection{Neighborhood Hyperparameters}
Our method for generating image neighborhoods introduces several hyperparameters: the type of metadata used, the size 
$m$ of each neighborhood, the max-rank $M$ for neighbors, and the tag-vocabulary size $\tau$.
Here we explore the influence of these hyperparameters on our model.

\begin{table}
\centering
\resizebox{0.49\textwidth}{!}{
\begin{tabular}{l||cc}
Method & mAP$_{L}$ & mAP$_{I}$ \\
\hline
\hline
CNN \cite{alex-2012} + logistic (visual-only) & $45.78 \scriptstyle{\pm 0.18}$ & $77.15 \scriptstyle{\pm 0.11}$ \\
\hline
Our model: visual neighbors & $ 47.45 \scriptstyle{\pm 0.19}$ & $78.56\scriptstyle{\pm 0.14}$ \\
Our model: group neighbors & $48.87 \scriptstyle{\pm 0.22}$ & $79.11 \scriptstyle{\pm 0.13}$ \\
Our model: set neighbors & $48.02 \scriptstyle{\pm 0.33}$ & $78.40 \scriptstyle{\pm 0.25}$ \\
Our model: tag neighbors & $\mathbf{52.78} \scriptstyle{\pm 0.34}$ & $\mathbf{80.34} \scriptstyle{\pm 0.07}$ \\
\hline
\end{tabular}
}
\vspace{-7pt}
\caption{Our model trained with different image neighborhoods \emph{vs} the visual-only model.}
\label{tab:our_models}
\vspace{-7pt}
\end{table}

\begin{figure}
\centering
\includegraphics[width=0.48\linewidth]{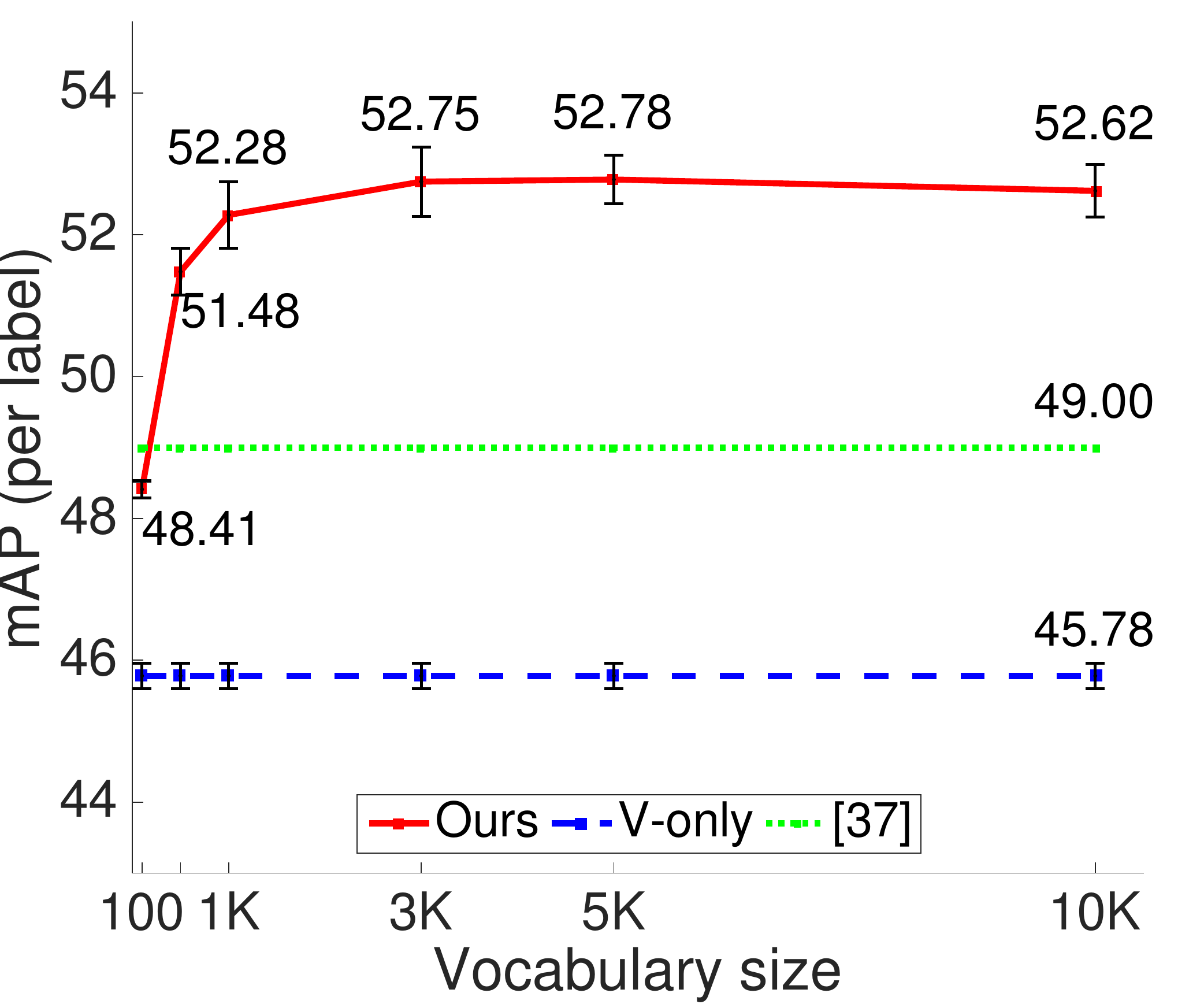}
\includegraphics[width=0.48\linewidth]{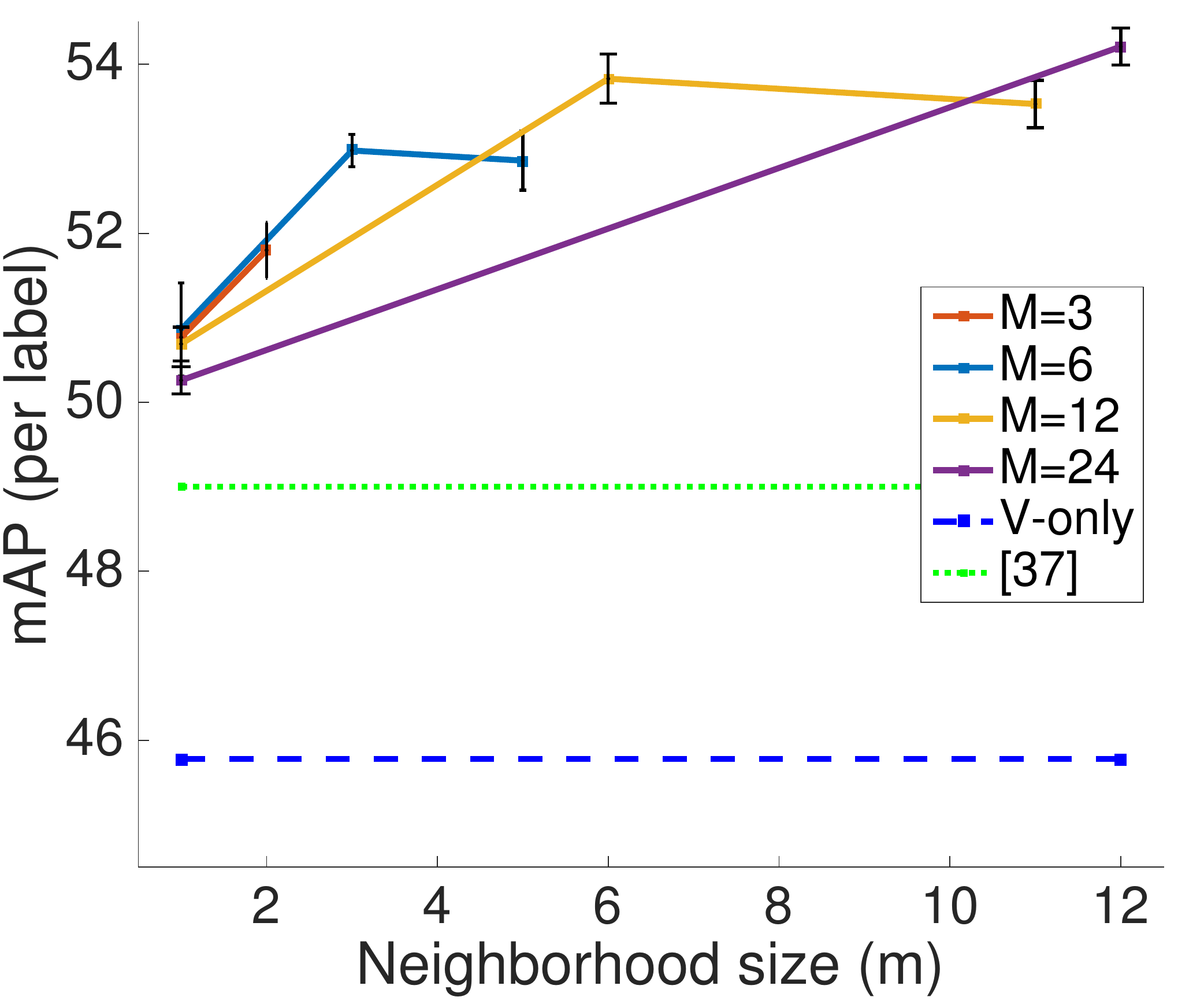}
\vspace{-1mm}
\caption{Performance of our model as we vary the neighborhood size $m$, max-rank $M$, and
  tag vocabulary size $\tau$. In all cases our model outperforms the baselines.\vspace{-4mm}}
\label{fig:neighborhood_params}
\end{figure}

\paragraph{Effects on performance.}\label{sec:hyperparameters}
The most important hyperparameter for generating neighborhoods is the type of data used.
We show in Table~\ref{tab:our_models} the performance of our model using different types of metadata: tags give the
highest performance, followed by groups and then sets. In all cases our model outperforms the visual-only baseline.
We also show the effect of using Euclidean distance of visual features to build neighborhoods (visual neighbors).
This setup slightly outperforms the visual-only baseline but is outperformed when using metadata, showing both the 
ability of our method to handle a variety of neighbor types, and the importance of image metadata.

To study the effects of the neighborhood size $m$, the max-rank $M$, and the tag vocabulary size $\tau$ we show
in Figure~\ref{fig:neighborhood_params} the performance of our model as we vary these hyperparameters.
Varying the max-rank $M$ gives the largest variation in performance, but in all cases we show improvements over
the visual-only baseline and the results of \cite{mcauley-2012}.

\paragraph{Label correlations.}
We can better interpret the influence of neighborhood hyperparameters by studying the correlations between the labels 
of images and their nearest neighbors. With strong correlations, visual evidence for a label among an image's 
neighbors is evidence that the image should have the same label; as such, our model should perform better when these 
correlations are stronger.

To this end, we plot in Figure~\ref{fig:prob_vs_k} the probability that the $k$th nearest neighbor of an image has a 
particular label given that the image itself has the label; on the same axis we show the baseline probability that a 
random image in the dataset has the label. This experiment shows that the nearest neighbors of images 
are indeed very likely to share labels with an image, and helps to explain the influence of various hyperparameters.
An image's labels are most highly correlated with its tag neighbors, followed by groups and then sets; this matches 
the results of Table~\ref{tab:our_models}. The flat shape of all curves in Figure~\ref{fig:prob_vs_k} suggests that 
the 20th nearest neighbor is nearly as informative as the 10th, suggesting that larger max-ranks $M$ may increase 
performance.

\subsection{Generalization Experiments}
One advantage of our model is that we only use metadata of images nonparametrically as a means to 
compute image neighborhoods. As a result, our model can easily cope with situations where different 
types of metadata are available during training and testing.

\paragraph{Vocabulary Generalization.}
Our best-performing model relies on user tags to generate image neighborhoods. In a real-world 
setting, the vocabulary of user tags may change over time: new tags may become popular, and older tags may fall into 
disuse. Any method that depends on user tags should be able to cope with these challenges.

Ideally, to test our model's resilience to changes in user tags over time, we would train the model using a snapshot 
of Flickr images at one point in time and test the model using a snapshot from a different point in time.

\begin{figure}
\centering
\includegraphics[width=.49\linewidth]{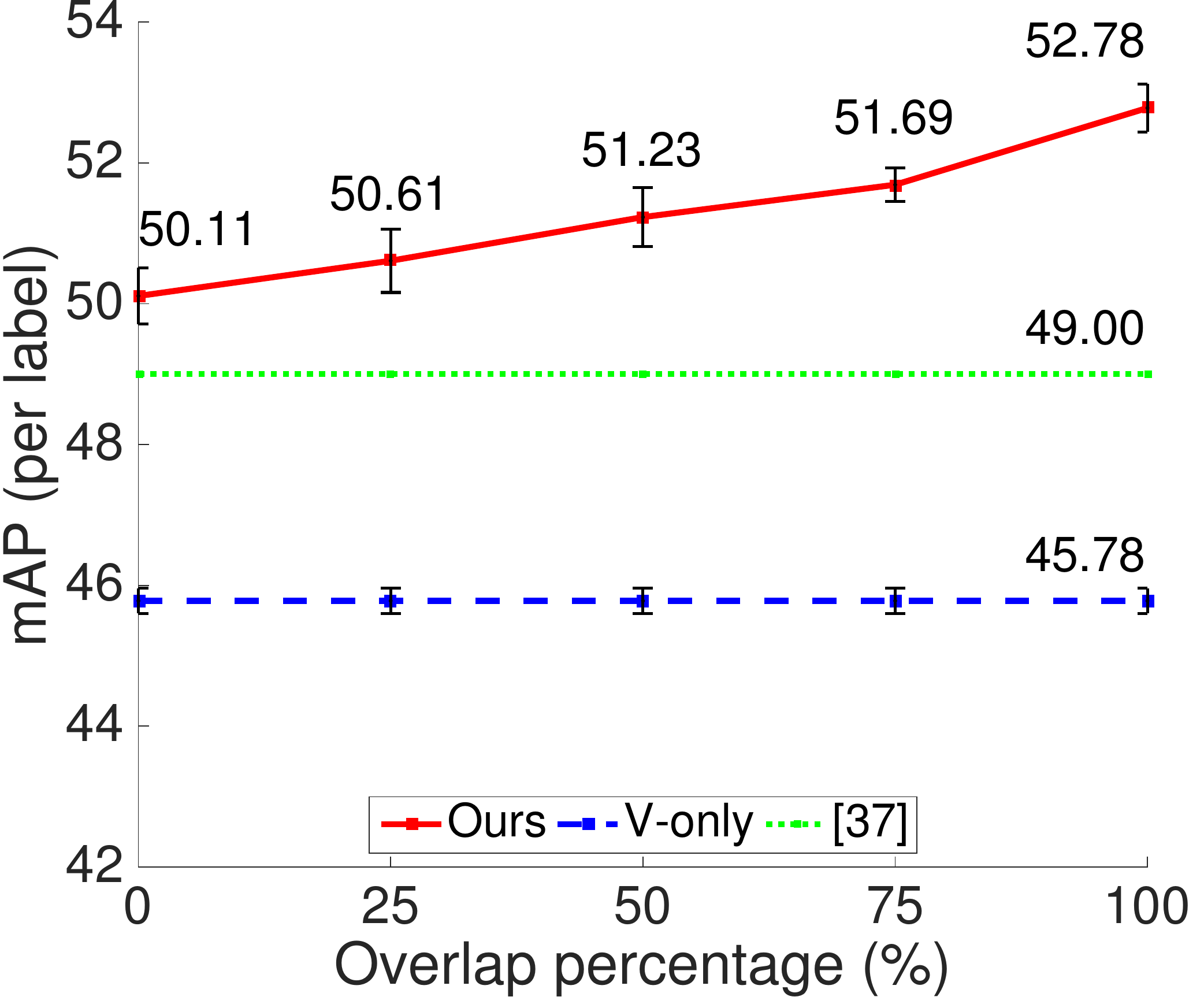}
\includegraphics[width=.49\linewidth]{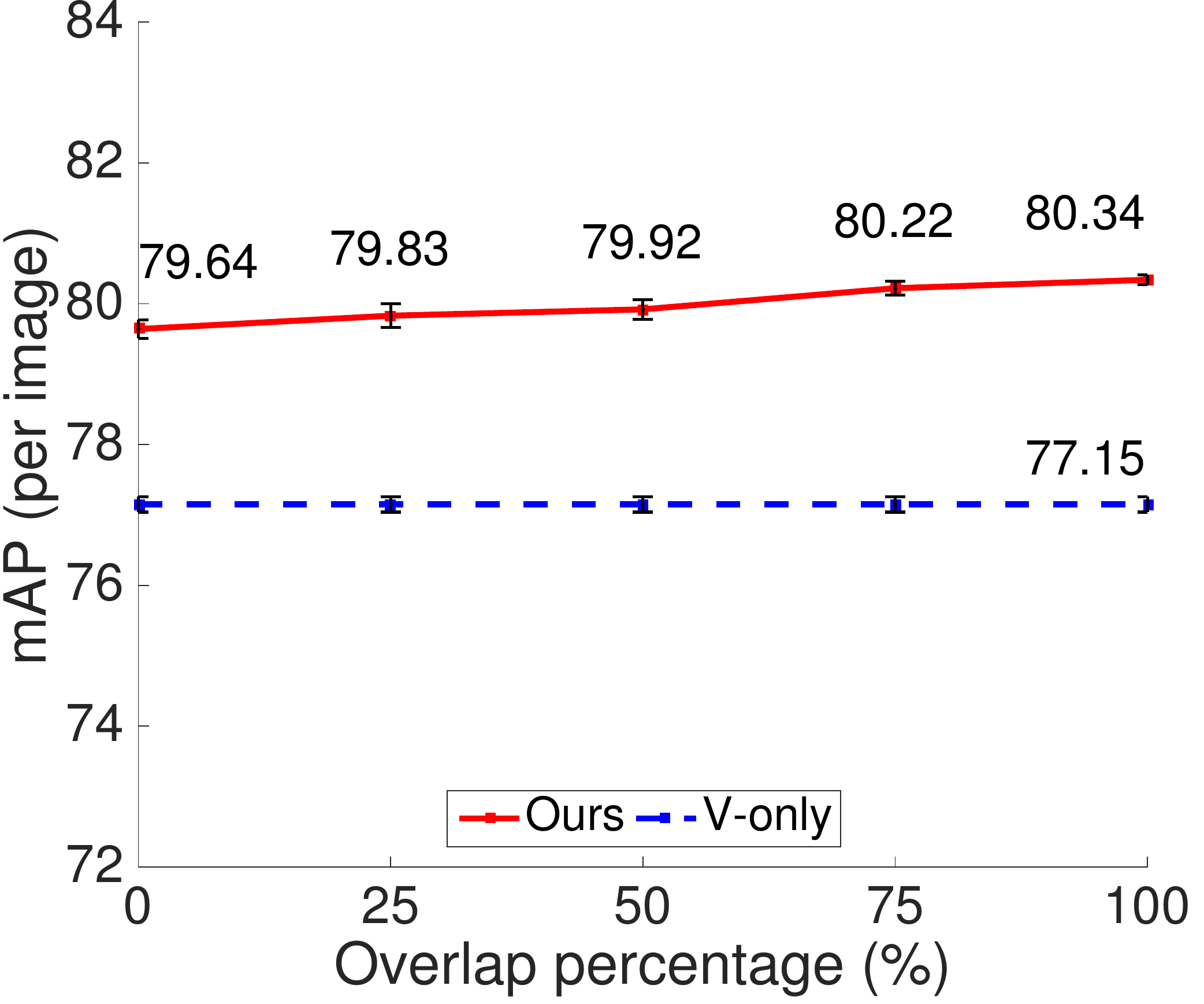}
\caption{
  Performance as we vary overlap between tag vocabularies used for training and testing.
  Our model gives strong results even in the case of disjoint vocabularies.}
\label{fig:overlap}
\end{figure}

Unfortunately we do not have access to this type of data. As a proxy to such an experiment, we instead randomly divide 
the 10K most commonly occurring user tags into two sets. During training we use the first set of user tags to generate 
neighborhoods, and use the second during testing.
We vary the degree to which the training tags and the testing tags overlap; with an overlap of 0\% there are no tags 
shared between training and testing, and an overlap of 100\% uses the same vocabulary of user tags for training and 
testing. Results are shown in Figure~\ref{fig:overlap}.

We see that the performance of our model degrades as we decrease the overlap between the training and testing tags; 
however even in the case of 0\% overlap our model is able to outperform both the visual-only model and 
\cite{mcauley-2012}.

\paragraph{Metadata Generalization.}
As a test of our model's ability to generalize across different types of metadata, we 
perform an experiment where we use different types of metadata during training and testing.
For example, we generate neighborhoods with tags during training and instead use sets during testing.
Table~\ref{tab:crosssig} shows the quantitative results of this experiment; in all cases our model
outperforms the visual-only baseline.
These results suggest that our model could be applied in cases where some types of metadata are 
unavailable during testing.

We can explain the results of this experiment by again examining Figure~\ref{fig:prob_vs_k}. When we train using one 
signal and test using another, our train and test data are no longer drawn from the same distribution, breaking 
one of the core assumptions of supervised learning. However the parametric portion of our model only 
views image metadata through the lens of nearest neighbors; Figure~\ref{fig:prob_vs_k} shows that 
changing the method of computing these neighbors does not drastically change the nature of the correlations between 
the labels of an image and its neighbors.

\begin{table}
\centering
\resizebox{0.48\textwidth}{!}{
\begin{tabular}{c|ccc}
\backslashbox{Train:}{Test:} & Tags & Sets & Groups\\
\hline
Tags & $52.78 \scriptstyle{\pm 0.34}$ & $47.12 \scriptstyle{\pm 0.35}$ & $48.14 \scriptstyle{\pm 0.33}$ \\
Sets & $52.21 \scriptstyle{\pm 0.29}$ & $48.02 \scriptstyle{\pm 0.33}$ & $48.49 \scriptstyle{\pm 0.16}$ \\
Groups & $50.32 \scriptstyle{\pm 0.28}$ & $47.82 \scriptstyle{\pm 0.24}$ & $48.87 \scriptstyle{\pm 0.22}$ \\
\hline
\end{tabular}
}
\vspace{-3pt}
\caption{Metadata generalization experiment. We use different types of metadata during training and 
  testing, and report mAP$_L$ for all possible pairs. All combinations outperform the visual-only model 
  ($45.78\scriptstyle{\pm 0.34}$).}
\vspace{-3mm}
\label{tab:crosssig}
\end{table}

\section{Conclusion}\vspace{-2mm}
We have introduced a framework that exploits image metadata to generate neighborhoods of images, 
and uses a strong parametric visual model based on deep convolutional neural networks to blend visual information
between an image and its neighbors. We use our model to achieve state-of-the-art performance for multilabel
image annotation on the NUS-WIDE dataset. We also show that our model gives impressive results even when it is
forced to generalize to new types of metadata at test time.

\vspace{-3mm}\paragraph{Acknowledgments.}
We thank J.~Leskovec, J.~Krause and O.~Russakovsky for helpful comments and discussions.
J.~Johnson is supported by a Magic Grant from The Brown Institute for Media Innovation and
L.~Ballan is supported by a Marie Curie Fellowship from the EU (623930).
We gratefully acknowledge the support of NVIDIA for their donation of GPUs and Yahoo for
their donation of cluster machines used in this research.

\vspace{-2mm}
{\scriptsize
\bibliographystyle{ieee}
\bibliography{seeingsocial}
}

\end{document}